\documentclass[10pt,twocolumn]{article}
\usepackage[margin=0.72in]{geometry}
\usepackage{graphicx}
\usepackage{amsmath,amssymb,amsfonts,bm}
\usepackage[numbers,sort&compress]{natbib}
\usepackage{hyperref}
\usepackage{authblk}
\usepackage{booktabs}
\usepackage{microtype}
\usepackage{adjustbox}
\usepackage{enumitem}
\setlist[itemize]{leftmargin=*,topsep=2pt,itemsep=1pt}
\title{Insect-inspired modular architectures as inductive biases for reinforcement learning}

\author[1]{A. E. Staples}
\affil[1]{Department of Mechanical Engineering, Virginia Tech, Blacksburg, Virginia 24061, USA}
\affil[ ]{\texttt{staplesa@vt.edu}}

\date{April 2026}

\begin{document}
\maketitle

\begin{abstract}
Most reinforcement-learning (RL) controllers used in continuous control are architecturally centralized: observations are compressed into a single latent state from which both value estimates and actions are produced. Biological control systems are often organized differently. Insects, in particular, coordinate navigation, heading stabilization, memory, and context-dependent action selection through distributed circuits rather than a single monolithic controller. Motivated by this contrast, we study an RL policy architecture that decomposes control into interacting modules for sensory encoding, heading representation, sparse associative memory, recurrent command generation, and local motor control, with a learned arbitration mechanism that allocates motor authority across modules. The model is evaluated on a two-dimensional navigation task that require simultaneous food seeking, obstacle avoidance, and predator escape. In a six-seed predator-navigation experiment trained with Proximal Policy Optimization (PPO) for 75 updates, the modular policy achieves the strongest final mean performance among the tested controllers, with final episodic return $-2798.8\pm964.4$ versus $-3778.0\pm628.1$ for a centralized gated recurrent unit (GRU) and $-4727.5\pm772.5$ for a centralized multilayer perceptron (MLP). The modular policy also attains the lowest final value loss and stable PPO optimization statistics while driving module-assignment entropy to $0.0457\pm0.0244$, indicating highly selective control allocation. These results suggest that distributed control can serve as a useful inductive bias for RL problems involving dynamically competing behavioral objectives.
\end{abstract}

\section{Introduction}
Centralized function approximators dominate modern reinforcement learning. Feedforward and recurrent neural-network policies have proved highly effective, but they typically compress perception, memory, and action selection into a single latent representation. This strategy is flexible, yet it does not reflect the distributed organization of many biological nervous systems. In insects, navigation and action selection rely on specialized structures such as the central complex, which is strongly implicated in heading and sensorimotor control, and the mushroom body, which supports sparse associative memory and reinforcement-driven learning \citep{Honkanen2019,Heinze2024,Seelig2015,Bennett2021}.

That biological contrast suggests a computational question: when control is posed as a dynamical problem in a partially observed continuous environment, can architectural decomposition itself act as a useful inductive bias? This question sits at the intersection of computational neuroscience, nonlinear dynamical control, and machine learning. It also connects to a long literature on modular and hierarchical RL, including temporal abstraction via options, feudal control, and mixtures of experts \citep{Sutton1999,Dayan1993,Jacobs1991}. The present work differs from those frameworks in emphasis. Rather than learning a hierarchy of abstract skills, it builds a controller around functionally distinct modules inspired by insect navigation and reinforcement circuits.

The proposed architecture separates (i) sensory encoding, (ii) a recurrent heading state reminiscent of central-complex ring-attractor models \citep{Seelig2015,Kim2017}, (iii) a sparse associative memory inspired by mushroom-body coding \citep{Ardin2016,Bennett2021}, (iv) a recurrent command center that emits behavioral mode probabilities and low-dimensional command signals, and (v) a bank of local controllers whose candidate actions are combined by a learned arbiter. The purpose is not biological realism per se, but to test whether distributed organization helps in tasks that require switching among competing objectives.

In a six-seed experiment reported here, across 75 PPO updates, the insect-inspired controller outperforms both centralized baselines in final mean episodic return, lowers the critic loss relative to both the MLP and GRU, and maintains PPO statistics comparable to the recurrent baseline while avoiding the optimization pathologies seen in the feedforward model. These results support the claim that distributed modular control appears systematically advantageous on this predator-navigation task.

\section{Task and learning setup}
The environment is a bounded two-dimensional world with continuous state and action spaces. The agent must navigate to a food target while avoiding static obstacles and, in the harder task, a moving predator. The observation vector in the predator setting has dimension $10$ and contains food direction, nearest-obstacle direction, heading features, food distance, nearest-obstacle distance, and predator direction. Actions are two-dimensional,
\begin{equation}
\mathbf{a}_t = (u_t,\omega_t),
\end{equation}
where $u_t$ is forward thrust and $\omega_t$ is turn rate. Heading evolves as
\begin{equation}
\theta_{t+1}=\theta_t+0.15\tanh(\omega_t),
\end{equation}
and position updates according to the heading direction with step size proportional to $\tanh(u_t)$.

The reward combines goal acquisition, survival pressure, and efficiency penalties. In the predator-navigation implementation, the instantaneous reward is
\begin{equation}
\begin{aligned}
r_t = {} & -0.01 + 10\,\mathbb{I}[\text{food reached}] - 25\,\mathbb{I}[d_{\rm pred}<2.5] \\
& - 2\max(0,3-d_{\rm pred}) - 10\,\mathbb{I}[\text{collision}] \\
& - 1.5\max(0,1.5-d_{\rm obs}),
\end{aligned}
\end{equation}
where $d_{\rm pred}$ is predator distance and $d_{\rm obs}$ is nearest-obstacle distance. Episodes end when the agent reaches food, collides, or times out at 256 steps.

Policies are trained with PPO \citep{Schulman2017}. Given generalized-advantage estimates $\hat A_t$, PPO maximizes the clipped surrogate
\begin{equation}
\mathcal{L}_{\rm PPO} = \mathbb{E}_t\Big[\min\big(r_t(\phi)\hat A_t,\; \mathrm{clip}(r_t(\phi),1-\epsilon,1+\epsilon)\hat A_t\big)\Big],
\end{equation}
with ratio $r_t(\phi)=\pi_\phi(a_t|o_t)/\pi_{\phi_{\rm old}}(a_t|o_t)$. The updated experiment uses six seeds, 75 PPO updates, 16 parallel environments, and rollout length 256, with $\gamma=0.99$, GAE parameter $\lambda=0.95$, clip coefficient $0.2$, learning rate $3\times10^{-4}$, and four PPO epochs per update. For the insect-inspired model, auxiliary terms encourage economical command signals, diffuse high-level mode probabilities, and sparse low-level arbitration.

\section{Distributed modular controller}
The policy is built from interacting modules with comparable overall parameter count to the centralized baselines: $476{,}063$ trainable parameters for the insect-inspired model, versus $438{,}021$ for the MLP and $464{,}133$ for the GRU.

The MLP baseline maps observations directly to policy and value outputs through a monolithic feedforward network. The GRU baseline first encodes the observation, then updates a single recurrent hidden state used by both the actor and critic.

\subsection{Sensory encoding}
The observation $\mathbf{o}_t\in\mathbb{R}^{10}$ is partitioned into vision-like features $\mathbf{o}^{(v)}_t\in\mathbb{R}^4$, proprioceptive features $\mathbf{o}^{(p)}_t\in\mathbb{R}^2$, and remaining task variables $\mathbf{o}^{(x)}_t\in\mathbb{R}^4$. Each stream is encoded by a separate multilayer perceptron,
\begin{equation}
\mathbf{v}_t=f_v(\mathbf{o}^{(v)}_t),\quad \mathbf{p}_t=f_p(\mathbf{o}^{(p)}_t),\quad \mathbf{x}_t=f_x(\mathbf{o}^{(x)}_t),
\end{equation}
then fused into a shared hidden representation
\begin{equation}
\mathbf{s}_t = f_{\rm fuse}([\mathbf{v}_t,\mathbf{p}_t,\mathbf{x}_t]).
\end{equation}
This is a mild but useful structural bias: directional obstacle cues, proprioception, and task variables are not forced to share the same first-stage features.

\subsection{Heading state and associative memory}
The heading module maintains a recurrent state $\mathbf{h}^{(\theta)}_t\in\mathbb{R}^{32}$ using sensory input, proprioception, the previous heading state, and a command-dependent turn signal,
\begin{equation}
\mathbf{h}^{(\theta)}_t = \mathrm{LN}\!\left[\tanh\!\left(W_s\![\mathbf{s}_t,\mathbf{o}^{(p)}_t] + \mathbf{h}^{(\theta)}_{t-1}K + W_c\mathbf{c}^{(0)}_t\right)\right],
\end{equation}
where $K$ is initialized as a ring-structured recurrent kernel. This gives the controller a persistent directional state loosely analogous to population codes studied in the insect central complex \citep{Seelig2015,Kim2017,Honkanen2019}.

Associative memory uses a sparse high-dimensional projection inspired by Kenyon-cell coding in the mushroom body,
\begin{equation}
\begin{aligned}
\mathbf{k}_t &= \mathrm{TopK}_{32}\big(\mathrm{ReLU}(W_m\mathbf{s}_t)\big), \\
\mathbf{m}_t &= \tanh\big(W_r\mathbf{k}_t + W_v[\mathbf{c}_t,\bm{\pi}_t]\big).
\end{aligned}
\end{equation}
where $\mathbf{k}_t\in\mathbb{R}^{512}$ is sparse, $\mathbf{m}_t\in\mathbb{R}^{64}$ is the memory readout, $\mathbf{c}_t\in\mathbb{R}^{16}$ is a low-dimensional command vector, and $\bm{\pi}_t\in\mathbb{R}^{6}$ is a probability distribution over behavioral modes. Sparse projection-and-readout is a natural computational abstraction of mushroom-body style expansion coding \citep{Ardin2016,Bennett2021,Webb2024}.

\subsection{Command center, local controllers, and arbiter}
A recurrent command center integrates fused sensory features, heading, and memory:
\begin{equation}
\mathbf{z}_t = \mathrm{GRUCell}\big([\mathbf{s}_t,\mathbf{h}^{(\theta)}_t,\mathbf{m}_t],\mathbf{z}_{t-1}\big).
\end{equation}
From $\mathbf{z}_t$ the model produces (i) mode logits $\bm{\ell}_t$, (ii) command vector $\mathbf{c}_t$, and (iii) critic value $V_t$,
\begin{equation}
\bm{\pi}_t=\mathrm{softmax}(W_{\pi}\mathbf{z}_t),\qquad \mathbf{c}_t=\tanh(W_c\mathbf{z}_t),\qquad V_t=W_V\mathbf{z}_t.
\end{equation}

Four local controllers then propose candidate motor actions: \textit{stabilize}, \textit{avoid}, \textit{approach}, and \textit{explore}. Each receives a context-specific local input $\mathbf{q}^{(j)}_t$ together with the shared command and mode signals,
\begin{equation}
\tilde{\mathbf{a}}^{(j)}_t = \tanh\,g_j([\mathbf{q}^{(j)}_t,\mathbf{c}_t,\bm{\pi}_t]), \qquad p^{(j)}_t = u_j([\mathbf{q}^{(j)}_t,\mathbf{c}_t,\bm{\pi}_t]).
\end{equation}
The arbiter converts priorities $p^{(j)}_t$ into weights
\begin{equation}
\alpha^{(j)}_t = \frac{\exp p^{(j)}_t}{\sum_k \exp p^{(k)}_t},
\end{equation}
and fuses local proposals as
\begin{equation}
\bar{\mathbf{a}}_t = \sum_j \alpha^{(j)}_t\tilde{\mathbf{a}}^{(j)}_t.
\end{equation}
A final Gaussian policy head yields mean action $\bm{\mu}_t$ and diagonal standard deviation $\bm{\sigma}$,
\begin{equation}
\pi(\mathbf{a}_t|\mathbf{o}_{\le t}) = \mathcal{N}(\bm{\mu}_t,\mathrm{diag}(\bm{\sigma}^2)).
\end{equation}

Training uses a total loss
\begin{equation}
\begin{aligned}
\mathcal{L} = {} & \mathcal{L}_{\rm PPO}
+ c_v \mathcal{L}_{\rm value}
- c_H \mathcal{H}(\pi)
+ \lambda_1 \|\mathbf{c}_t\|_1 \\
& - \lambda_m H(\bm{\pi}_t)
+ \lambda_a H(\bm{\alpha}_t) \, ,
\end{aligned}
\end{equation}
where the last three terms correspond to the code-level auxiliary penalties and entropies. Because the arbiter entropy enters with positive coefficient in the loss, minimizing $\mathcal{L}$ encourages sparse module selection.

\begin{figure*}[t]
\centering
\includegraphics[width=\linewidth]{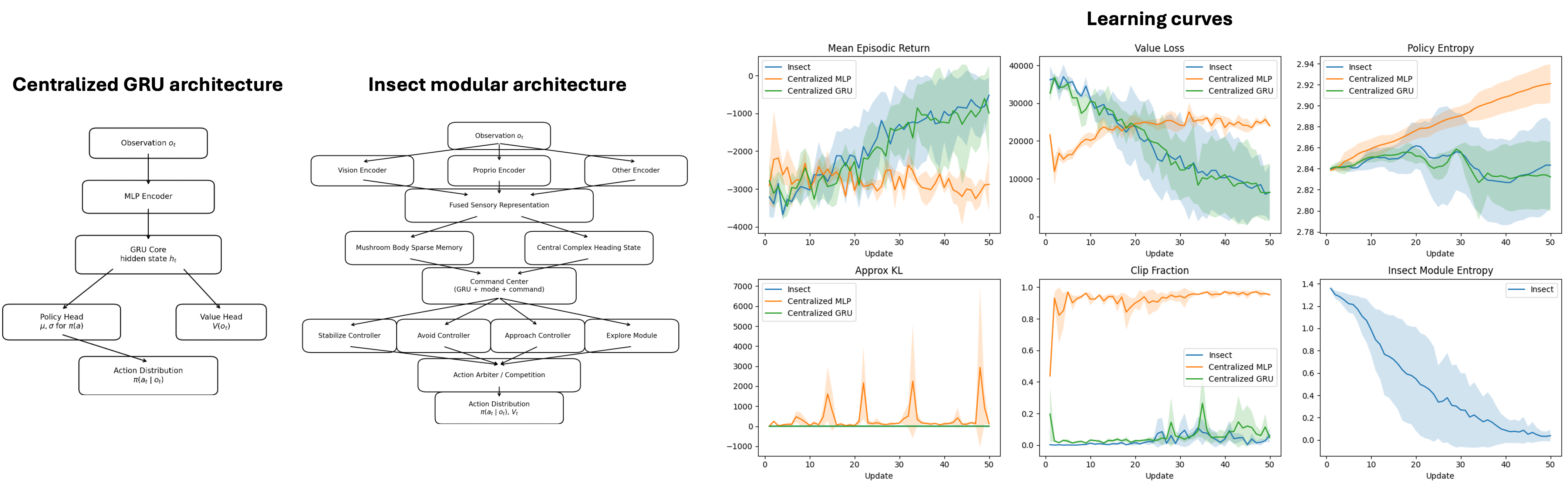}
\caption{\textbf{Modular control for behavioral switching.} \emph{Left}: a conventional recurrent RL policy uses a centralized latent state. \emph{Center}: the insect-inspired architecture distributes control across specialized modules with a learned arbitration mechanism. \emph{Right}: in the three-seed predator-navigation experiment, the modular architecture achieved the strongest final mean performance among the tested policies, while module-assignment entropy decreased steadily during training, consistent with increasingly selective internal allocation of control. Final returns were $-515.0212\pm471.0264$ (insect-inspired), $-986.5884\pm1244.2115$ (GRU), and $-2880.4218\pm717.9300$ (MLP). Relative to the GRU and MLP baselines, the insect-inspired model improved the final mean episodic return by $47.8\%$ and $82.1\%$, respectively, while reducing return variability by $62.1\%$ and $34.4\%$.}
\label{fig:main}
\end{figure*}

\section{Results}
Three- and six-seed experiments were conducted and provide clear evidence in favor of the modular controller. Figure~\ref{fig:main} summarizes the model architecture and the results of the three-seed experiment. Table~\ref{tab:predator_results} summarizes the final metrics after 75 PPO updates in the six-seed experiment. The insect-inspired architecture achieved the best final mean episodic return, $-2798.8\pm964.4$, outperforming both the centralized GRU, $-3778.0\pm628.1$, and the centralized MLP, $-4727.5\pm772.5$. Because higher episodic return is better, this corresponds to an improvement of roughly $979$ return units over the GRU and $1929$ over the MLP. The modular controller also yielded the lowest final value loss, $45842.3\pm24958.6$, compared with $60992.2\pm7377.6$ for the GRU and $66463.1\pm5284.3$ for the MLP.

\begin{table*}[t]
\centering
\caption{Final predator-navigation performance after 75 PPO updates over 6 random seeds (mean $\pm$ standard deviation). Higher episodic return is better, with values closer to zero indicating stronger performance. Lower value loss, clip fraction, KL, command magnitude, and module entropy indicate more stable or more selective control. Percentage columns report relative improvement of the insect-inspired model over each baseline, using the appropriate direction for each metric; negative percentages indicate worse performance on that metric.}
\vspace{0.6em}
\label{tab:predator_results}
\scriptsize
\begin{adjustbox}{max width=\textwidth}
\begin{tabular}{lccccc}
\toprule
Metric & Insect-inspired & Centralized MLP & Centralized GRU & Relative improvement vs.\ MLP & Relative improvement vs.\ GRU \\
\midrule
Mean episodic return & $-2799 \pm 964$ & $-4728 \pm 772$ & $-3778 \pm 628$ & $40.8\%$ & $25.9\%$ \\
Return across seed std.\ dev. & $964$ & $772$ & $628$ & $-24.9\%$ & $-53.5\%$ \\
Value loss & $45842 \pm 24959$ & $66463 \pm 5284$ & $60992 \pm 7378$ & $31.0\%$ & $24.8\%$ \\
Policy entropy & $2.901 \pm 0.040$ & $2.900 \pm 0.012$ & $2.838 \pm 0.030$ & --- & --- \\
Approximate KL divergence & $0.0030 \pm 0.0013$ & $2383.1 \pm 4392.0$ & $0.0030 \pm 0.0015$ & --- & --- \\
Clip fraction & $0.023 \pm 0.016$ & $0.974 \pm 0.011$ & $0.024 \pm 0.017$ & $97.7\%$ & $3.8\%$ \\
Command magnitude ($L_1$) & $0.496 \pm 0.066$ & --- & --- & --- & --- \\
Mode entropy & $1.743 \pm 0.043$ & --- & --- & --- & --- \\
Module entropy & $0.046 \pm 0.024$ & --- & --- & --- & --- \\
\bottomrule
\end{tabular}
\end{adjustbox}
\end{table*}

Optimization diagnostics help clarify the interpretation of the results. The insect-inspired controller and the GRU have nearly identical approximate KL divergence ($0.0030\pm0.0013$ and $0.0030\pm0.0015$, respectively) and similarly low clip fractions ($0.0228\pm0.0160$ and $0.0237\pm0.0165$), indicating that both operate in a stable PPO regime. The MLP, by contrast, is clearly unstable: its final approximate KL is $2383.1\pm4392.0$ and its clip fraction is $0.9742\pm0.0106$, showing that nearly all updates are clipped. This strongly suggests that the feedforward controller is mismatched to the partially observed predator-navigation task.

The modular-controller internal diagnostic values are also informative. The final module entropy is $0.0457\pm0.0244$, indicating highly selective allocation of motor authority, while the mode entropy remains substantially larger at $1.7433\pm0.0431$. This separation between high-level behavioral mixture and low-level decisive arbitration is one of the most interesting outcomes of the experiment: the policy appears to retain a distributed internal representation of behavioral regime even while producing sparse final control selection.

Figure~\ref{fig:main} summarizes the architecture and the learning behavior. The decline in module entropy reflects increasingly selective module allocation during training, and the revised final-return comparison confirms that this selectivity is associated with the strongest average task performance among the tested architectures.


\section{Discussion and conclusion}
In both three- and six-seed studies, the insect-inspired modular model was the best-performing architecture on average, and it achieved this without relying on unstable optimization. The key comparison is with the centralized GRU. Both models possess a recurrent state and both train stably under PPO, yet the insect-inspired architecture achieves a substantially better final mean return and lower value loss. This suggests that the advantage is not simply due to recurrence, but to the way recurrence is structured and distributed across specialized subsystems.

Three conclusions follow. First, explicit architectural decomposition can improve control in tasks with dynamically competing objectives. The predator task is difficult not because of raw perceptual complexity, but because the agent must repeatedly renegotiate which objective dominates. Separating heading state, context-sensitive memory, command generation, and local control proposals appears to be a better inductive bias than forcing every computation through one recurrent bottleneck.

Second, the benefit of the architecture is most naturally interpreted in dynamical-systems terms. The controller does not merely have multiple subnetworks. It factorizes the policy into interacting state variables with different computational roles: a directional recurrent state, a sparse memory code, a low-dimensional command manifold, and a simplex of arbitration weights. That factorization likely changes the geometry of learning by reducing interference between incompatible behavioral drives. From a computational-physics perspective, the model is therefore interesting not only as a bio-inspired policy, but as a structured nonlinear controller for agent--environment dynamics.

Third, these results clarify both the promise and the remaining challenges. The low final module entropy indicates that the architecture successfully learns decisive control allocation, but highly selective arbitration does not by itself guarantee ideal context switching. Future work should therefore test mechanisms that promote context-dependent specialization more directly, for example reflex pathways for predator escape, diversity penalties on controller usage, or conditional arbitration objectives that reward different modules for dominating in different environmental regimes. At the same time, this average performance advantage comes with higher across-seed variability in episodic return, indicating that robustness remains an important target for future work.

Overall, these experiments support the conclusion that insect-inspired distributed control is not merely viable, but appears systematically advantageous on the predator-navigation benchmark tested here at the tested training scale. The broader lesson is that biologically motivated structure can be useful in RL because it suggests mathematically meaningful decompositions of perception, memory, and control.

\section*{Acknowledgements}
The author thanks Kirubaharan Natarajan for assistance with computational experiments.

\end{document}